\documentclass{article}

    \usepackage[preprint]{neurips_2025}

\usepackage[utf8]{inputenc} 
\usepackage[T1]{fontenc}    
\usepackage{url}            
\usepackage[pagebackref=true,breaklinks=true,letterpaper=true,colorlinks,bookmarks=false,citecolor=blue]{hyperref}
\usepackage[table]{xcolor}
\usepackage{booktabs}       
\usepackage{multirow}      
\usepackage{makecell}
\usepackage{pifont}
\usepackage{todonotes}
\usepackage{bm}
\usepackage{graphicx}
\usepackage{amsfonts}       
\usepackage{nicefrac}       
\usepackage{microtype}      
\usepackage{graphicx}
\usepackage{subfigure}
\usepackage{array}
\usepackage{wrapfig}
\usepackage{amsmath}
\usepackage{enumitem}
\newcommand{\etal}{{\emph{et al.}}}
\newcommand{\ie}{{\emph{i.e.}}}
\newcommand{\eg}{{\emph{e.g.}}}

\title{SAILViT: Towards Robust and Generalizable Visual Backbones for MLLMs via Gradual Feature Refinement}

\author{%
  Weijie Yin$^{1}$\footnotemark[2]   $\quad$  Dingkang Yang$^{1,2}$\footnotemark[2]
  $\quad$  Hongyuan Dong$^{1}$ $\quad$ Zijian Kang$^{1}$ \\
  $\quad$ \textbf{Jiacong Wang}$^{1}$ 
  $\quad$ \textbf{Xiao Liang}$^{1}$\footnotemark[1]  $\quad$ \textbf{Chao Feng}$^{1}$\footnotemark[4]
  $\quad$ \textbf{Jiao Ran}$^{1}$
  \\
   \small $^{1}$ByteDance Inc.  \\
 \small $^{2}$College of Intelligent Robotics and Advanced 
Manufacturing, Fudan University  \\
  \small \texttt{$\{$yinweijie, yangdingkang$\}$} \\
   \small  \texttt{$\{$liangxiao.ilx, chaofeng.zz, ranjiao$\}$@bytedance.com}  \\
}

\begin{document}

\maketitle
\renewcommand{\thefootnote}{\fnsymbol{footnote}} 
\footnotetext[2]{Equal first contributions.} 
\footnotetext[1]{Project lead.}
\footnotetext[4]{Corresponding author.}  

\begin{abstract}
Vision Transformers (ViTs) are essential as foundation backbones in establishing the visual comprehension capabilities of Multimodal Large Language Models (MLLMs). Although most ViTs achieve impressive performance through image-text pair-based contrastive learning or self-supervised mechanisms, they struggle to engage in connector-based co-training directly with LLMs due to potential parameter initialization conflicts and modality semantic gaps. To address the above challenges, this paper proposes SAILViT, a gradual feature learning-enhanced ViT for facilitating MLLMs to break through performance bottlenecks in complex multimodal interactions. SAILViT achieves coarse-to-fine-grained feature alignment and world knowledge infusion with gradual feature refinement, which better serves target training demands. We perform thorough empirical analyses to confirm the powerful robustness and generalizability of SAILViT across different dimensions, including parameter sizes, model architectures, training strategies, and data scales. Equipped with SAILViT, existing MLLMs show significant and consistent performance improvements on the OpenCompass benchmark across extensive downstream tasks.
SAILViT series models are released at~\url{https://huggingface.co/BytedanceDouyinContent}.
\end{abstract}

\vspace{-5pt}
\section{Introduction}
\vspace{-5pt}

As the backbone architecture for learning general visual representations, 
research on vision foundation models has received growing attention
over time~\cite{jia2021scaling,karpathy2015deep,he2017mask,feichtenhofer2019slowfast,carion2020end,fan2025scaling}. Convolutional neural networks 
inspired by residual connections~\cite{he2016deep} initially proved that scaling up the depth and size of the models with sufficient training and data held promise for achieving superior performances~\cite{dai2017deformable,hu2018squeeze,xie2017aggregated}. 
In recent years, the emergence of Vision Transformer (ViT) structures~\cite{dosovitskiy2020image} has laid a solid foundation for pursuing more powerful visual feature expressiveness. ViTs and their variants~\cite{dosovitskiy2020image,wang2022pvt,radford2021learning,dehghani2023scaling,chen2022vision,cai2022reversible} have been widely applied to remarkable Multimodal Large Language Models (MLLMs)~\cite{wang2024world,wang2025vgr,lei2025scalability,wei2025learning,dong2025adalrs,dong2024benchmarking,xiao2024seeing}, which show potential in diverse multimodal tasks.

With the emergence of Large Language Models (LLMs)~\cite{ChatGPT}, mainstream encoders on the visual side are attached with LLMs through customized connectors (\eg, Q-Former~\cite{li2023blip} and Multilayer Perceptrons (MLPs)~\cite{liu2023visual,liu2024improved}), enabling MLLMs~\cite{wang2024qwen2,alayrac2022flamingo,zhu2023minigpt,lai2024lisa,zhang2023video} to achieve natural language reasoning with versatile visual perception capabilities.
Existing MLLMs~\cite{dong2025scalable,bai2023qwenvl,wang2024qwen2,alayrac2022flamingo,chen2024expanding,chen2024internvl} accommodate visual encoders that are typically pre-trained from two patterns: multimodal contrastive learning and autoregressive modeling. The former~\cite{tschannen2025siglip,zhai2023sigmoid,chen2024internvl} utilizes large-scale image-text pairs to capture discriminative features, but struggles to fulfill the demands of sophisticated spatial perception tasks due to scaling limitations and noise interference~\cite{fan2025scaling}. Despite the latter~\cite{el2024scalable,fini2024multimodal} realizing more intensive supervision than contrastive objectives via generative approaches, 
\begin{wrapfigure}{r}{0.6\textwidth}
\vspace{-10pt}
  \centering
  \setlength{\abovecaptionskip}{0pt}
  \setlength{\belowcaptionskip}{-8pt}
  \includegraphics[width=0.5\textwidth]{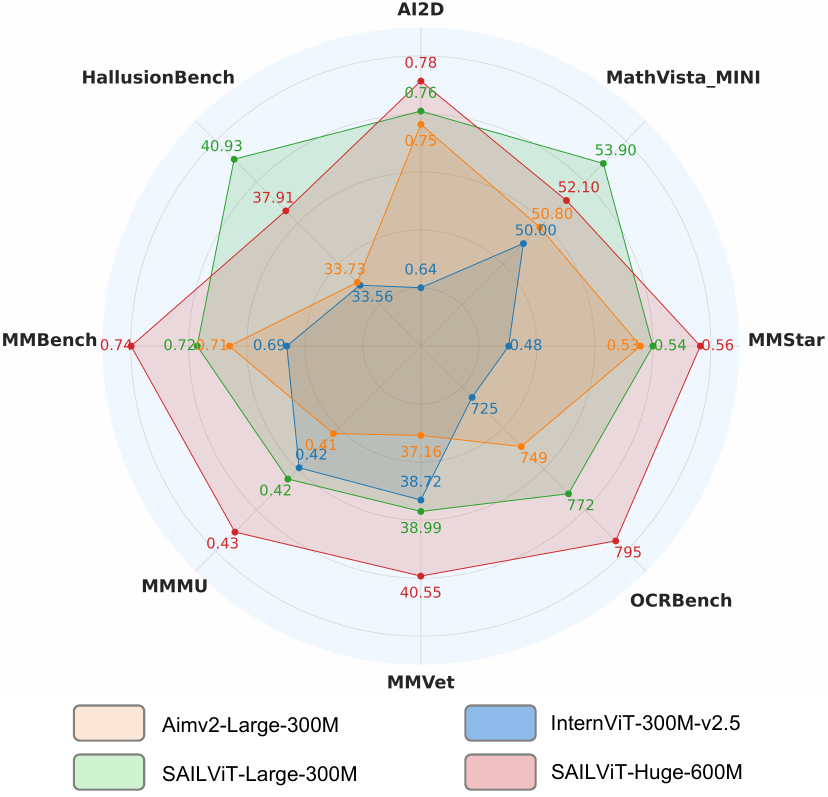}
  \caption{Comparison results of Qwen2.5-1.5B~\cite{yang2024qwen2} paired with different visual backbones across diverse tasks on the popular OpenCompass benchmark. Our SAILViT series is superior to AIMv2~\cite{fini2024multimodal} and InternViT~\cite{chen2024internvl} by large margins.
  } 
  \label{radar}
\end{wrapfigure}
it is difficult to ensure parameter alignment and consistent enhancement with LLMs for objective optimization~\cite{fini2024multimodal}.
Furthermore, existing efforts in pre-training procedures usually focus on constructing basic perception expertise~\cite{kakaobrain2022coyo700m,fang2023data} and lack the understanding of multimodal data containing comprehensive world knowledge. The above shortcomings induce performance bottlenecks caused by suboptimal visual backbones.

To yield more robust and generalized visual representations, this paper proposes SAILViT, a series of vision foundation models with all-around performance improvements based on gradual feature refinement.
The core philosophy is to mitigate parameter initialization conflicts and modality semantic discrepancies among components by progressive learning before MLLMs perform target training. Specifically, the gradual refinement strategy initially utilizes the MLP-driven connector as a bridge to accomplish coarse-grained modality alignment warm-up, aiming to build the fundamental visual perception capabilities. Subsequently, we instruct the trainable ViT to mine informative visual representations in enriched data scale and complexity to fulfill fine-grained modality alignment.
After that, all component parameters are turned on to facilitate the visual backbone in high-quality multimodal instructions to incorporate comprehensive world knowledge and narrow parameter distribution gaps with the LLMs, yielding our SAILViT that better satisfies the subsequent target training needs.
The main contributions of this work are as follows:

\begin{itemize}[leftmargin=*]
\item  We release SAILViT, the versatile visual backbone that eliminates multimodal perception performance bottlenecks on sophisticated visual understanding for constructing more powerful MLLMs.

\item We carefully design a training pipeline for gradual feature refinement, which accommodates different vision foundation models in a hierarchical and flexible optimization pattern, and achieves cross-modal alignment and integrated reinforcement of visual representations.

\item We prove the robustness and generalizability of SAILViT via multi-dimensional analyses, including different parameter sizes, model architectures, training strategies, and data scales. As Figure~\ref{radar} shows, SAILViT-Large outperforms the same-sized visual backbones by large margins on diverse tasks, while its Huge version assists MLLMs to achieve significant performance gains.

\end{itemize}

\vspace{-5pt}
\section{Related Work}
\vspace{-5pt}

\subsection{Vision Foundation Models}
\vspace{-3pt}

Visual encoders play an indispensable role as the core components of MLLMs to perceive visual information. Current State-of-the-Art (SOTA) methods commonly adopt the ViT-based architectures, which show noticeable differences in their pretraining strategies.
For instance, the CLIP~\cite{radford2021learning} and SigLIP~\cite{zhai2023sigmoid,tschannen2025siglip} series are pre-trained based on image-text pair data for contrastive learning. InternViT~\cite{chen2024internvl} introduces autoregressive loss on the basis of image-text pairs to further optimize the modality alignment.
Meanwhile, AIMv2~\cite{fini2024multimodal} combines generative and self-supervised optimization objectives for joint training.
In addition, some prominent works (\eg, LLaVA-HR~\cite{luo2024feast}, Mini-Gemini~\cite{li2024mini}, and DeepSeek-VL~\cite{lu2024deepseek}) explore the fusion modeling paradigm of multiple backbones by combining low-resolution ViTs (\eg, CLIP-ViT~\cite{radford2021learning}/SigLIP-L~\cite{zhai2023sigmoid}) with high-resolution convolutional networks (\eg, ConvNeXt~\cite{woo2023convnext}) to achieve multiscale feature extraction.
Further, Fan~\etal~\cite{fan2025scaling} analyze the effect of different pretraining losses on ViT through a series of data and model parameter scaling experiments, with the aim of exploring visual representations that are better aligned with LLMs.
Unfortunately, existing methods fail to model sophisticated multimodal world knowledge when integrated into MLLMs since pretraining data are mainly limited to basic perception-based data such as image-text pairs and OCR. In contrast, our proposed SAILViT adopts an incremental three-stage feature refinement strategy to achieve cross-modal alignment and rich world knowledge infusion from coarse to fine grains. Experimental results show that SAILViT-Large/Huge outperforms existing methods by up to 12.3\% on all 8 datasets of the OpenCompass benchmark, exhibiting the potential for vision foundation models.

\vspace{-3pt}
\subsection{Multimodal Large Language Model Evolution}
\vspace{-3pt}

With the excellent achievements of large-scale pre-trained models in practical applications~\cite{yang2024pediatricsgpt,yang2025improving,ChatGPT}, MLLMs~\cite{lai2024lisa,zhu2023minigpt,chen2024internvl,chen2024expanding,alayrac2022flamingo,huang2024mini,bai2025qwen2,zhu2025internvl3,team2025kimi,lu2024ovis,li2024llava} have evolved rapidly and gradually converged to the ``visual encoder + connector + LLM'' paradigm.
The mainstream methods generally adopt the instruction-tuned LLM (\eg, Qwen2.5-Instruct) as the base model, typically including the Qwen2.5-VL~\cite{bai2025qwen2} and Ovis~\cite{lu2024ovis} series.
Emerging works such as Kimi-VL~\cite{team2025kimi} and InternVL3~\cite{zhu2025internvl3} explore native multimodal pretraining pathways based on Moonlight-base~\cite{liu2025muon} and Qwen2.5-base~\cite{yang2024qwen2}, respectively. Among them, the connector usually appears as a nexus linking the vision and language modality feature spaces, whose general candidates are dominated by Q-Former~\cite{li2023blip} and MLP~\cite{liu2023visual,liu2024improved}.
Recent alternatives have gradually converged on MLP due to the considerable reduction in computational complexity while maintaining performance. To handle long sequential visual inputs, the dominant scheme utilizes the pixel shuffle technique~\cite{chen2024far}, \ie, 4-fold feature compression is achieved by merging 2×2 neighboring patches, which effectively balances computational efficiency and feature preservation.
Despite impressive advances through connector-based modality projections, current MLLMs typically suffer from parameter initialization conflicts and inherent semantic discrepancies across components~\cite{dong2025scalable}, leading to performance bottlenecks in the target training procedure.

\begin{table}[t]
\caption{Detailed statistical information on the training data at different training stages of SAILViT.}
\setlength{\tabcolsep}{-3pt}
\resizebox{\textwidth}{!}{%
\begin{tabular}{cccc}
\toprule
\textbf{Training Stage} & \textbf{Data Type} & \textbf{Data Source} & \textbf{Data Size} \\ \midrule
\multirow{2}{*}{\makecell[c]{Coarse-grained \\ Alignment}} & Caption & SAIL-Caption~\cite{dong2025scalable} & 4.9M \\
 & OCR & IDL-WDS~\cite{biten2022ocr} & 3.1M \\ \midrule
\multirow{3}{*}{\makecell[c]{Fine-grained \\ Alignment}} & Caption & SAIL-Caption & 6.7M \\
 & OCR & IDL-WDS, DocStruct~\cite{wang2020docstruct} & 3.3M \\
 & Video Caption & ShareGPTVideo~\cite{zhang2024direct}, Charades~\cite{sigurdsson2016hollywood}, Mementos~\cite{wang2024mementoscomprehensivebenchmarkmultimodal} & 1M \\ \midrule
\multirow{7}{*}{\makecell[c]{World Knowledge \\ Infusion}} & Caption & SAIL-Caption~\cite{dong2025scalable}, Molmo~\cite{deitke2024molmopixmoopenweights} & 3.1M\\
 & OCR & Docmatix~\cite{laurençon2024buildingbetterunderstandingvisionlanguage}, DocVQA~\cite{mathew2021docvqadatasetvqadocument}, DocStruct~\cite{wang2020docstruct}, Molmo & 9.6M \\
 & OpenQA & ShareGPT4V~\cite{chen2023sharegpt4vimprovinglargemultimodal}, ShareGPT4o~\cite{cui2025comprehensive} & 7.1M \\
 & Text & Openhermes 2.5~\cite{OpenHermes2.5}, Magpie~\cite{xu2024magpiealignmentdatasynthesis} & 4.8M \\
 & Math & MathQA~\cite{amini2019mathqainterpretablemathword}, MathV360k~\cite{shi2024mathllavabootstrappingmathematicalreasoning}, GeoQA+~\cite{cao2022augmented} & 1.5M \\
 & ShortQA & OKVQA~\cite{marino2019okvqavisualquestionanswering}, VQA v2~\cite{goyal2017makingvvqamatter}, ST-VQA~\cite{biten2022ocr} & 0.9M \\
 & Mix Data & Infinity-MM~\cite{gu2025infinitymmscalingmultimodalperformance}, Cauldron~\cite{laurençon2024matters}, Cambrian~\cite{tong2024cambrian1fullyopenvisioncentric}, LLaVA-1.5~\cite{liu2024improved}, LLaVA-OneVision~\cite{li2024llava} & 9M \\ \bottomrule
\end{tabular}%
}
\vspace{-5pt}
\label{tab1}
\end{table}

\vspace{-5pt}
\section{Methodology}
\vspace{-5pt}
This section describes the proposed gradual feature refinement training procedure for SAILViT. 
As shown in Figure~\ref{arc}, the core philosophy is to progressively inculcate hierarchical multimodal knowledge for the vision foundation model through three training stages in order to enhance visual representations and accomplish a seamless modality-semantic transition with LLMs. Formulaically, we optimize the standard autoregressive objective. Given an arbitrary training sample $\bm{x} = (x_0, x_1, ..., x_L)$ with the token length of $L$, we define each token can be expressed as a textual token embedding, a visual embedding, or a video patch embedding for brevity. The objective supervision followed in the three stages are as follows:
\begin{equation}
\mathcal{L}(\theta) = 
-   \sum_{\substack{i = 2\\x_{i}\in\text{Text}}}^{L} \mathrm{log}\,p(x_{i} \mid x_1, ..., x_{i-1};\theta ),
\end{equation}
where $\theta$ is the model parameter. $x_{i}\in\text{Text}$ represents the loss calculation is constrained to the text tokens for the gradient propagation.
Table~\ref{tab1} shows the statistical information of the training data.

\vspace{-3pt}
\subsection{Coarse-grained Modality Alignment}
\vspace{-3pt}

At this stage, we formulate randomly initialized the MLP projector as a connector to accomplish the pre-training alignment. The connector is trained while keeping the vision and language models frozen. To facilitate the initial bridging of visual and linguistic representations, we use the 8M basic perception training corpus as shown in Table~\ref{tab1}, including 4.9M high-quality caption data sampled from SAIL-Caption~\cite{dong2025scalable} and 3.1M OCR data from IDL-WDS~\cite{biten2022ocr}.
In one training epoch, the learning rate and batch size are set to 2e-4 and 1,920, respectively, which facilitates the rapid adaptation of the connector to feature mappings in different modality semantic spaces. Coarse-grained feature alignment benefits the connector by providing a suitable optimization warm-up for subsequent unlocking of more trainable parameters and improving training stability.

\vspace{-3pt}
\subsection{Fine-grained Modality Alignment}
\vspace{-3pt}
 \begin{figure}[t]
  \centering
  \includegraphics[width=\linewidth]{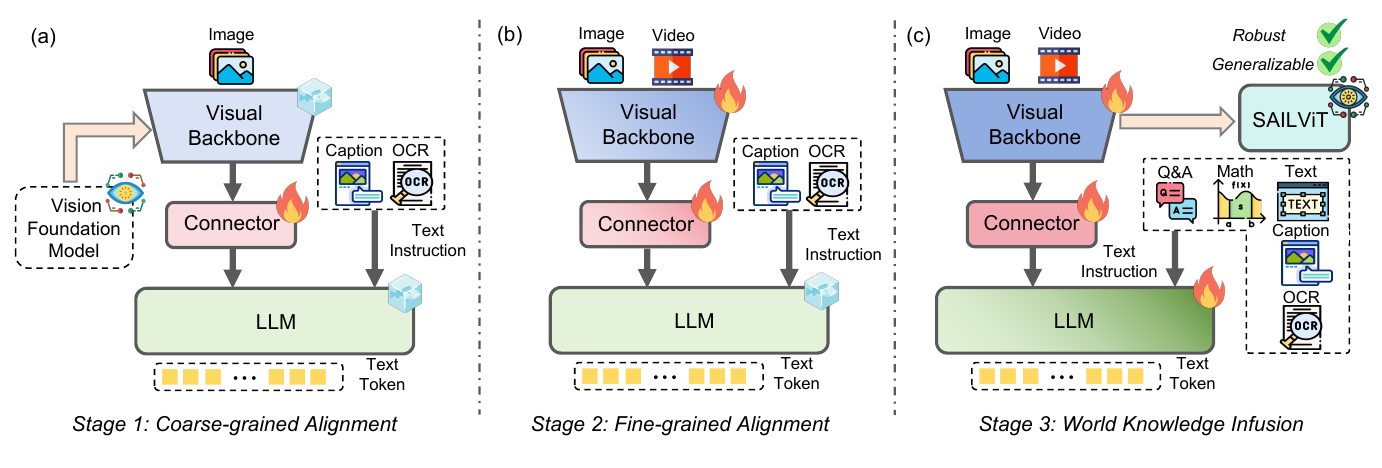}
  \caption{Illustration of the training pipeline for SAILViT by gradual feature refinement. (a) In the coarse-grained alignment stage, we optimize the MLP connector to achieve initial adaptation across the modality feature space while improving subsequent training stability. (b) In the fine-grained alignment stage, we simultaneously train the visual backbone and connector in a multi-task manner to enhance the expressiveness and richness of visual element semantics. (c) All parameters are unlocked during the world knowledge infusion stage to perform joint training. We utilize multimodal instruction data to strengthen the multifaceted and comprehensive extraction capability for visual representations, resulting in well-performing SAILViT.}
  \label{arc}
  \vspace{-0.35cm}
\end{figure}
The fine-grained alignment aims to enhance the extraction of diverse visual features by unlocking the visual backbone and connector for simultaneous optimization. To accommodate the trainable parameters, we introduce additional video caption data (\ie, ShareGPTVideo~\cite{zhang2024direct}, Charades~\cite{sigurdsson2016hollywood}, Mementos~\cite{wang2024mementoscomprehensivebenchmarkmultimodal}) to perform multi-task pre-training at this stage. This can better help the visual backbone capture the semantics of visual elements more closely aligned to real-world distributions than the image captions.
Regarding data reuse, 6.7M resources from the SAIL-Caption subset are utilized to expand the diversity of visual Q\&A distributions. We also supplement the OCR data from DocStruct~\cite{wang2020docstruct}. In this case, we adjust the learning rate and batch size to 2e-5 and 512 in one epoch to prevent performance degradation due to potential catastrophic forgetting and model collapse~\cite{shumailov2023curse}.

\vspace{-3pt}
\subsection{World Knowledge Infusion}
\vspace{-3pt}
As shown in Figure~\ref{arc}\textcolor{red}{c}, we concurrently unlock the visual backbone, the connector, and the LLM to perform training under one epoch with a learning rate of 1e-5 and a batch size of 512. The full-parameter pattern allows the visual backbone to more sufficiently invoke the intensive knowledge from the pre-training period and strengthen the logical reasoning efficacy during the dynamic adaptation of feature semantics.
To this end, a large number of mixed data sources (\ie, Q\&A~\cite{gu2025infinitymmscalingmultimodalperformance,laurençon2024matters,tong2024cambrian1fullyopenvisioncentric,liu2024improved,li2024llava,chen2023sharegpt4vimprovinglargemultimodal,cui2025comprehensive,marino2019okvqavisualquestionanswering,goyal2017makingvvqamatter,biten2022ocr}, Math~\cite{amini2019mathqainterpretablemathword,shi2024mathllavabootstrappingmathematicalreasoning,cao2022augmented}, and OCR~\cite{laurençon2024buildingbetterunderstandingvisionlanguage,mathew2021docvqadatasetvqadocument}) covering different disciplines, topics, and domains are introduced to facilitate model learning for hierarchical and multifaceted visual representations. We also consider the inclusion of plain text data~\cite{OpenHermes2.5,xu2024magpiealignmentdatasynthesis} to maintain the original generative capacity of LLMs.

\vspace{-5pt}
\section{Experiments}
\vspace{-5pt}
In this section, we first introduce the implementation details and benchmark evaluation. 
Then, we systematically prove the effectiveness, generalizability, and robustness of SAILViT in different dimensions. In particular, we demonstrate the transferability of the proposed three-stage training procedure to other visual backbones.

\vspace{-3pt}
\subsection{Implementation Details}
\vspace{-3pt}
The SAILViT series is constructed based on AIMv2~\cite{fini2024multimodal} and consists of two versions with 300M and 600M parameters to produce our \textbf{SAILViT-Large} and \textbf{SAILViT-Huge}.
SAILViT accepts image inputs of fixed resolution (specifically 448×448). For larger-resolution images, we adopt a tiling strategy with a maximum of 8 sub-images. With a patch size of 14, 1024 visual tokens are extracted per fixed-sized image. To model larger-resolution images, we employ a pixel shuffle~\cite{chen2024far} strategy to compress the original visual token count to 256.
We integrate the SAILViT series with Qwen2.5-1.5B~\cite{yang2024qwen2} using a 2-layer MLP, and model training is conducted on the PyTorch platform. Within the training pipeline, the first stage requires 1,664 NPU hours and the second stage consumes 2,200 NPU hours.
During the supervised fine-tuning stage, we consume 10,304 NPU hours. All trainings are implemented on top of multiple Ascend 910B2.
To alleviate memory overhead during training, we utilize DeepSpeed ZeRO3~\cite{rajbhandari2020zeromemoryoptimizationstraining} and gradient checkpointing strategies. For network optimization, we adopt the AdamW~\cite{loshchilov2019decoupledweightdecayregularization} optimizer and select bf16 data precision.

\vspace{-3pt}
\subsection{Benchmark Evaluation}
\label{sec:eval}
\vspace{-3pt}
\textbf{Evaluation}.
We evaluate visual backbones on a broader range of VQA tasks based on MLLMs that adopt a two-stage training paradigm. Unless otherwise specified, the LLM integrated with the evaluated ViT is Qwen2.5-1.5B. Specifically, we first use the 8M image caption and OCR data from SAIL-Caption~\cite{dong2025scalable} and IDL-WDS~\cite{biten2022ocr} to train the connector, aiming to align the feature spaces of visual and language modalities. This stage employs a learning rate of 2e-4 and a batch size of 1,920. During the second stage, we conduct full-parameter fine-tuning using 3M question-answer (Q\&A) pairs from Infinity-MM~\cite{gu2025infinitymmscalingmultimodalperformance}, utilizing a learning rate of 1e-5 and a batch size of 512. 
Finally, we evaluate different MLLMs using the customized version of the VLMEvalKit~\cite{duan2024vlmevalkit}.

\textbf{Benchmark}.
We implement extensive experiments on the well-known OpenCompass~\cite{2023opencompass} benchmark and the challenging OpenSource benchmark to achieve systematic evaluations of MLLMs based on different ViTs.
OpenCompass has 8 evaluation datasets across diverse multimodal tasks, including AI2D~\cite{kembhavi2016diagramworthdozenimages}, HallusionBench~\cite{guan2024hallusionbenchadvanceddiagnosticsuite}, MMMU~\cite{yue2024mmmumassivemultidisciplinemultimodal}, MMStar~\cite{chen2024rightwayevaluatinglarge}, MMVet~\cite{yu2024mmvetevaluatinglargemultimodal}, OCRBench~\cite{Liu_2024}, MMBench~\cite{liu2024mmbenchmultimodalmodelallaround}, and MathVista\_MINI~\cite{lu2024mathvistaevaluatingmathematicalreasoning}.
In addition, we extend an OpenSource benchmark that incorporates 11 additional public datasets based on OpenCompass. These rich datasets include ChartQA~\cite{masry2022chartqabenchmarkquestionanswering}, DocVQA~\cite{masry2022chartqabenchmarkquestionanswering}, InfoVQA~\cite{mathew2021infographicvqa}, TextQA~\cite{singh2019vqamodelsread}, LLaVABench~\cite{liu2024improved}, MME~\cite{fu2024mmecomprehensiveevaluationbenchmark}, OCRVQA~\cite{mishraICDAR19}, POPE~\cite{li2023evaluatingobjecthallucinationlarge}, RealWorldQA~\cite{realworldqa}, SEEDBench~\cite{li2023seed}, and ScienceQA~\cite{lu2022learnexplainmultimodalreasoning}. Extensive datasets ensure robust assessments across diverse visual question answering tasks, including natural image/video analysis, OCR-based document understanding, complex reasoning scenarios demanding domain knowledge integration and beyond.

\textbf{Model Zoo}.
We compare a diverse set of available LLMs to enhance the robustness of SAILViT's evaluation. Specifically, InternLM2.5-1.8B (Chat)~\cite{cai2024internlm2} is a lightweight yet efficient conversational LLM optimized for dialogue scenarios and serves as the LLM component of InternVL2.5-2B~\cite{chen2024expanding}. Qwen2.5-1.5B/7B (Instruct)~\cite{yang2024qwen2} represent popular LLM choices among MLLM researchers: trained on 18 trillion tokens, these models support 128k context windows and demonstrate excellence in multilingual understanding, code generation, and mathematical reasoning. Qwen3-0.6B/1.7B/8B~\cite{qwen3} are part of a newly released LLM series that leverage a three-stage pretraining strategy and ``big-teaches-small'' distillation to optimize performance across diverse scenarios. We select visual backbones of comparable scale from the current top-performing MLLMs on the OpenCompass benchmark as baseline models for comparison with SAILViT, specifically choosing AIMv2-Large/Huge~\cite{fini2024multimodal} and InternViT-300M-v2.5~\cite{chen2024expanding}. The former (300M/600M) are ViTs trained by Apple using image-text pair data, with the training loss being a combination of image pixel MSE loss and text autoregressive loss. The latter is a ViT released by the Shanghai AI Lab, trained solely on image-text pair data.

\vspace{-3pt}
\subsection{Comparison Results on OpenCompass Benchmark}
\vspace{-3pt}

\begin{table}[t]
\caption{
Comparison results between SAILViT and other open-source ViTs on the OpenCompass benchmark when connecting to different LLM series with different parameters. The best results are bolded on the quantitative metrics.}
\resizebox{\textwidth}{!}{%
\renewcommand{\arraystretch}{1.3}
\setlength{\tabcolsep}{3pt}
\begin{tabular}{c|cc|c|cccccccc}
\toprule
\multicolumn{3}{c|}{Base Models} & \multirow{2}{*}{Avg.} & \multirow{2}{*}{\makecell[c]{AI2D \\ (test)}} & \multirow{2}{*}{HallusionBench} & \multirow{2}{*}{\makecell[c]{MMBench \\ (val)}} & \multirow{2}{*}{\makecell[c]{MMMU \\ (val)}} & \multirow{2}{*}{MMVet} & \multirow{2}{*}{OCRBench} & \multirow{2}{*}{MMStar} & \multirow{2}{*}{\makecell[c]{MathVista \\ (testmini)}} \\ \cline{1-3}
\multicolumn{1}{c}{LLM Series} & \multicolumn{2}{c|}{Visual Backbones Series} &  &  &  &  &  &  &  &  &  \\ 
\midrule
\rowcolor{gray!15} \multicolumn{12}{c}{\textit{InternLM2.5 Series}} \\ 
\midrule
\multirow{5}{*}{InternLM2.5-1.8B} & InternViT-300M-v2.5 & \multicolumn{1}{l|}{\multirow{3}{*}{300M}} & 49.9 & 69.27 & 33.35 & 65.98 & 35.11 & 29.59 & 713 & 47.80 & 47.1 \\
 & AIMv2-Large & \multicolumn{1}{l|}{} & 51.2 & 71.53 & 33.02 & 66.45 & 35.33 & 32.34 & \textbf{729} & 49.93 & 48.0\\
 & \textbf{SAILViT-Large} & \multicolumn{1}{l|}{} & \textbf{52.4} & \textbf{73.09} & \textbf{35.32} & \textbf{67.80} & \textbf{36.78} & \textbf{34.40} & 716 & \textbf{50.80} & \textbf{49.3} \\
 \cline{2-12}
 & AIMv2-Huge & \multicolumn{1}{l|}{\multirow{2}{*}{600M}} & 51.9 & 72.44 & 29.84 & 68.34 & 33.11 & 34.40 & \textbf{759} & 51.47 & 49.7 \\
 & \textbf{SAILViT-Huge} & \multicolumn{1}{l|}{} & \textbf{54.4} & \textbf{73.19} & \textbf{37.62} & \textbf{70.16} & \textbf{36.89} & \textbf{37.16} & 757 & \textbf{53.20} & \textbf{51.2} \\ 
\midrule
\rowcolor{gray!15} \multicolumn{12}{c}{\textit{Qwen2.5 Series}} \\ 
\midrule
\multirow{5}{*}{Qwen2.5-1.5B} & InternViT-300M-v2.5 & \multicolumn{1}{l|}{\multirow{3}{*}{300M}} & 52.2 & 64.02 & 33.56 & 69.16 & 41.56 & 38.72 & 725 & 48.33 & 50.0 \\
 & AIMv2-Large & \multicolumn{1}{l|}{} & 54.6 & 75.29 & 33.73 & 70.94 & 40.56 & 37.16 & 749 & 53.33 & 50.8 \\
 & \textbf{SAILViT-Large} & \multicolumn{1}{l|}{} & \textbf{56.9} & \textbf{76.20} & \textbf{40.93} & \textbf{71.94} & \textbf{41.89} & \textbf{38.99} & \textbf{772} & \textbf{53.89} & \textbf{53.9} \\
 \cline{2-12}
 & AIMv2-Huge & \multicolumn{1}{l|}{\multirow{2}{*}{600M}} & 56.3 & 77.66 & 35.44 & 72.33 & 42.56 & 39.72 & 769 & 54.13 & 51.4 \\
 & \textbf{SAILViT-Huge} & \multicolumn{1}{l|}{} & \textbf{57.7} & \textbf{78.27} & \textbf{37.91} & \textbf{73.99} & \textbf{43.44} & \textbf{40.55} & \textbf{795} & \textbf{55.60} & \textbf{52.1} \\ \midrule
\multirow{5}{*}{Qwen2.5-7B} & InternViT-300M-v2.5 & \multicolumn{1}{l|}{\multirow{3}{*}{300M}} & 62.1 & 81.28 & 44.82 & 77.67 & 49.44 & 43.90 & 784 & 59.27 & 62.2 \\
 & AIMv2-Large & \multicolumn{1}{l|}{} & 63.7 & 81.74 & 45.36 & 77.78 & 49.00 & 48.07 & \textbf{828} & 60.80 & 64.0 \\
 & \textbf{SAILViT-Large} & \multicolumn{1}{l|}{} & \textbf{64.5} & \textbf{82.12} & \textbf{45.63} & \textbf{78.95} & \textbf{51.67} & \textbf{49.50} & 805 & \textbf{60.87} & \textbf{67.1} \\
 \cline{2-12}
 & AIMv2-Huge & \multicolumn{1}{l|}{\multirow{2}{*}{600M}} & 64.2 & 81.44 & 44.04 & \textbf{80.30} & \textbf{50.78} & 46.10 & 815 & 62.33 & \textbf{67.4} \\
 & \textbf{SAILViT-Huge} & \multicolumn{1}{l|}{} & \textbf{65.2} & \textbf{83.00} & \textbf{48.65} & 79.64 & 50.33 & \textbf{49.22} & \textbf{833} & \textbf{62.60} & 65.2 \\ 
\midrule
\rowcolor{gray!15} \multicolumn{12}{c}{\textit{Qwen3 Series}} \\ 
\midrule
\multirow{2}{*}{Qwen3-0.6B} & AIMv2-Large & \multicolumn{1}{l|}{\multirow{2}{*}{300M}} & 51.7 & \textbf{71.37} & 38.92 & \textbf{65.52} & 35.11 & 33.07 & 703 & 49.80 & 49.2 \\
 & \textbf{SAILViT-Large} & \multicolumn{1}{l|}{} & \textbf{52.9} & 71.05 & \textbf{41.05} & 64.86 & \textbf{36.67} & \textbf{34.77} & \textbf{741} & \textbf{50.93} & \textbf{49.4} \\ \midrule
\multirow{4}{*}{Qwen3-1.7B} & AIMv2-Large & \multicolumn{1}{l|}{\multirow{2}{*}{300M}} & 56.3 & 77.49 & 39.76 & 71.28 & \textbf{42.78} & 39.17 & 751 & 53.40 & 51.4 \\
 & \textbf{SAILViT-Large} & \multicolumn{1}{l|}{} & \textbf{58.1} & \textbf{78.82} & \textbf{41.84} & \textbf{71.44} & 42.56 & \textbf{43.21} & \textbf{790} & \textbf{54.80} & \textbf{52.8} \\
 \cline{2-12}
 & AIMv2-Huge & \multicolumn{1}{l|}{\multirow{2}{*}{600M}} & 57.7 & 79.24 & 40.44 & 71.09 & 42.00 & 43.07 & 781 & 54.80 & 52.9 \\
 & \textbf{SAILViT-Huge} & \multicolumn{1}{l|}{} & \textbf{59.4} & \textbf{79.89} & \textbf{41.23} & \textbf{74.19} & \textbf{43.11} & \textbf{44.31} & \textbf{806} & \textbf{56.93} & \textbf{54.9} \\ \midrule
\multirow{2}{*}{Qwen3-8B} & AIMv2-Huge & \multicolumn{1}{l|}{\multirow{2}{*}{600M}} & 66.0 & 83.35 & 43.50 & 80.92 & \textbf{52.78} & \textbf{50.05} & 839 & \textbf{65.60} & 67.8 \\
 & \textbf{SAILViT-Huge} & \multicolumn{1}{l|}{} & \textbf{66.6} & \textbf{84.17} & \textbf{48.23} & \textbf{81.89} & 51.78 & 46.83 & \textbf{857} & 65.13 & \textbf{69.4} \\ \bottomrule
\end{tabular}
}
\label{opencompass}
\vspace{-3pt}
\end{table}

\textbf{Effectiveness of Pairing with Different LLMs}. Table~\ref{opencompass} shows the results of SAILViT in comparison with other visual backbones connected with LLMs on OpenCompass. We find that SAILViT has significant performance gains on different families of LLMs. As an example in SAILViT-Lagre, our model obtains average relative improvements of 2.3\%, 5.5\%, and 5.4\% on InternLM2.5, Qwen2.5, and Qwen3 series, respectively. Consistent gains are also observed in the Huge version. These results confirm that building more robust visual backbones can break the performance bottleneck of MLLMs.

\textbf{Cross-task Performance Analysis}.
Further, we investigate the impact of SAILViT on different tasks. On the HallusionBench task, SAILViT-Huge can significantly increase the performance from 29.84\% to 37.62\% after pairing it with InterLM2.5-1.8B compared to AIMv2-Huge.
Meanwhile, SAILViT-Large boosts performance by 7.2\% after plugging into Qwen2.5-1.5B, taking a new SOTA.
This observation suggests that SAILViT can mitigate the multimodal hallucination interferences of MLLMs by extracting more fine-grained and holistic visual semantics to improve the factuality of the model responses.
In addition, different versions of our SAILViT-based models achieve a majority of benefits in multidisciplinary multimodal Q\&A and logical reasoning~\cite{chen2024rightwayevaluatinglarge,yue2024mmmumassivemultidisciplinemultimodal}.

\textbf{Model Generalizability Assessment}. We explore the effect of SAILViT on different sizes of LLMs. For the Qwen2.5 series, we first observe that SAILViT has consistent boosts on both 1.5B and 7B model scales, with SAILViT-Large/Huge outperforming the 7B version on the 1.5B version.
Similarly, we find in the Qwen3 series that SAILViT-lage/huge improves better on 1.7B than its 0.6B and 8B counterparts. A plausible explanation for the above phenomena is that SAILViT has a better proximity to feature alignment for models of the same scale as the LLMs that are involved in gradual learning during training. In summary, SAILViT has broad generalizability across different sizes of LLMs.

\textbf{Model Robustness Assessment}. The results in Table~\ref{opencompass} also demonstrate that the performance gains of SAILViT can be adapted to different series of LLMs. Specifically, Although our visual backbone is combined with the adaptation training performed on Qwen2.5-1.5B, SAILViT shows significant and consistent performance improvements on both the InternLM series as well as the Qwen3 series of models. Noticeably, SAILViT-Large always outperforms AIMv2-Huge on average performance across the different LLM series, verifying that the designed gradual feature
refinement strategy is able to break through the scaling law caused by parameter constraints to a certain extent.
In the Appendix~\ref{sec: futher verification SAILViT robustness}, we further confirm the robustness of SAILViT by introducing the gradual alignment training stage for ViTs into the existing ViT evaluation regimes.

\begin{table}[t]
\caption{Comparison results between SAILViT and other open-source ViTs on the OpenSource benchmark when connecting to different LLM series with different parameters. The best results are bolded on the quantitative metrics. OpenSource includes a total of 19 evaluation datasets.}
\renewcommand{\arraystretch}{1.2}
\setlength{\tabcolsep}{4pt}
\resizebox{\textwidth}{!}{%
\begin{tabular}{cccccccc}
\toprule
\multicolumn{2}{c}{\multirow{2}{*}{Visual Backbone Series}} & \multicolumn{6}{c}{LLM Series} \\ \cline{3-8} 
\multicolumn{2}{c}{} & \multicolumn{1}{c|}{InternLM2.5-1.8B} & Qwen2.5-1.5B & \multicolumn{1}{c|}{Qwen2.5-7B} & Qwen3-0.6B & Qwen3-1.7B & Qwen3-8B \\ \midrule
InternViT-300M-v2.5 & \multirow{3}{*}{300M} & \multicolumn{1}{c|}{56.74} & 59.09 & \multicolumn{1}{c|}{67.17} & - & - & - \\
AIMv2-Large &  & \multicolumn{1}{c|}{57.34} & 60.05 & \multicolumn{1}{c|}{68.98} & 57.45 & 62.03 & - \\
\textbf{SAILViT-Large} &  & \multicolumn{1}{c|}{\textbf{59.09}} & \textbf{62.46} & \multicolumn{1}{c|}{\textbf{70.04}} & \textbf{59.64} & \textbf{64.47} & - \\ \midrule
AIMv2-Huge & \multirow{2}{*}{600M} & \multicolumn{1}{c|}{58.38} & 61.07 & \multicolumn{1}{c|}{69.71} & - & 63.19 & 69.60 \\
\textbf{SAILViT-Huge} &  & \multicolumn{1}{c|}{\textbf{60.48}} & \textbf{62.72} & \multicolumn{1}{c|}{\textbf{70.90}} & - & \textbf{65.86} & \textbf{70.58} \\ \bottomrule
\end{tabular}%
}
\label{opensource}
\vspace{-5pt}
\end{table}

\vspace{-3pt}
\subsection{Comparison Results on OpenSource Benchmark}
\vspace{-3pt}

The OpenSource benchmark is more challenging in terms of breadth and depth compared to OpenCompass because it introduces 11 additional open-source test sets. We mainly report the average results in Table~\ref{opensource} for intuitive comparisons.
The Appendix~\ref{sec:a} shows the fine-grained comparison results on the OpenSource benchmark.
Obviously, SAILViT steadily improves the overall performance of MLLMs when paired with different sizes and different series of LLMs. For instance, SAILViT-Large surpasses AIMv2-Large and InternViT-300M-v2.5 by at least 2\% gain advantage when paired with the latest Qwen3 series models. Moreover, SAILViT-Huge enhances the performance of the MLLMs with a gain interval of 0.98\%-2.67\% when dealing with multidisciplinary multimodal tasks targeting the understanding of complex vision-language semantics. The above observations further confirm the powerful generalizability and robustness of SAILViT.

\begin{figure}[t]
  \centering
  \includegraphics[width=0.9\linewidth]{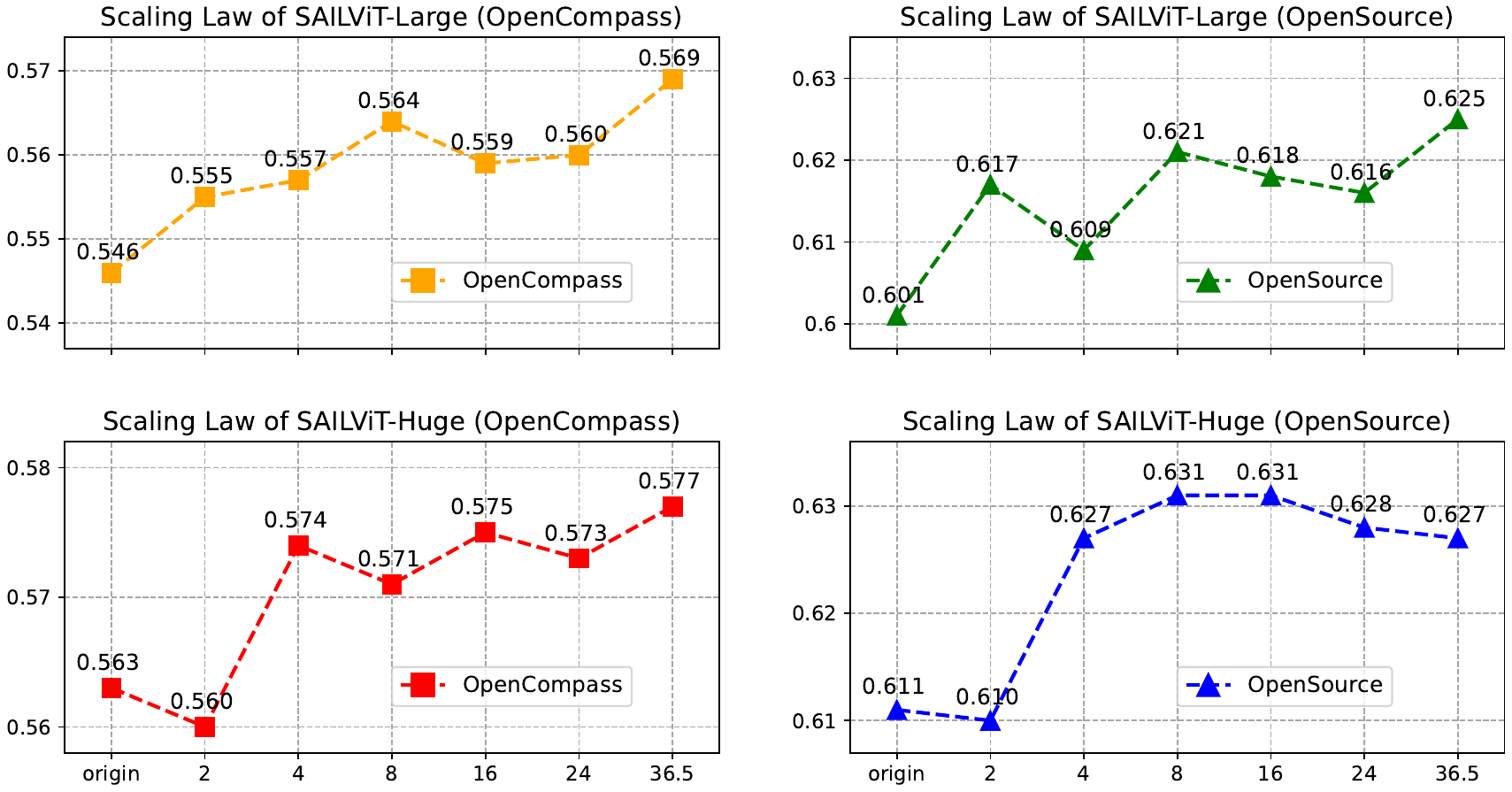}
  \caption{Illustration of the scaling law phenomenon analysis. The overall performances of MLLMs paired with SAILViT and Qwen2.5-1.5B on OpenCompass and OpenSource benchmarks show increasing trends as the amount of data in the world knowledge infusion stage increases. ``origin'' means the results of the MLLMs equipped with AIMv2-Large and AIMv2-Huge.}
  \label{fig:scaling}
  \vspace{-6pt}
\end{figure}

\vspace{-5pt}
\subsection{Scaling Law Analysis}
\vspace{-5pt}

To verify the training data scale effect, we construct the same distribution but different magnitudes of data having rich world knowledge in the knowledge infusion stage of SAILViT. In this case, we obtain several sets of Large and Huge versions of our model and connect them into MLLMs for training and evaluation, respectively.
As shown in Figure~\ref{fig:scaling}, SAILViT-Large/Huge exhibits obvious scaling law properties on both OpenCompass and OpenSource metrics. In other words, as the data size in world knowledge infusion training increases, the visual features extracted by SAILViT are more compatible with the representation space of LLMs, leading to higher metrics of MLLMs on downstream tasks. Additionally, experiments conducted with Qwen2.5-0.5B futher validated this view, as detailed in the Appendix~\ref{sec: scaling law under 0d5B}.

\begin{table}[t]
\centering
\caption{Ablation results on SAILViT in the world knowledge infusion stage. We consider three dimensions: different modules with trainable parameters, the scale of the integrated LLMs, and the adopted training strategy (step-by-step multiple SFT stages \textit{vs}. merging all SFT stages into one).}
\renewcommand{\arraystretch}{1}
\setlength{\tabcolsep}{5pt}
\resizebox{0.8\textwidth}{!}{%
\begin{tabular}{cc|ccccc}
\toprule
\multicolumn{2}{c}{Ablation Configuration} & Baseline & Setting 1 & Setting 2 & Setting 3 & Best Setting \\ \midrule
\multirow{3}{*}{Trainable Parameters} & ViT & - & \ding{51} & \ding{51} & \ding{51} & \ding{51} \\
 & Adapter & - & \ding{51} & \ding{51} & \ding{51} & \ding{51} \\
 & LLM & - & - & \ding{51} & \ding{51} & \ding{51} \\ \midrule
\multirow{2}{*}{LLM} & Qwen2.5-0.5B & - & - & \ding{51} & - & - \\
 & Qwen2.5-1.5B & - & \ding{51} & - & \ding{51} & \ding{51} \\ \midrule
\multirow{2}{*}{SFT Strategy} & Step by Step & - & - & - & \ding{51} & - \\
 & AIO & - & \ding{51} & \ding{51} & - & \ding{51} \\ \midrule
\multicolumn{2}{c}{OpenSource} & 60.10 & 62.45 & 61.10 & 61.80 & \textbf{62.46} \\
\multicolumn{2}{c}{OpenCompass} & 54.60 & 55.30 & 55.40 & 55.80 & \textbf{56.90} \\ \bottomrule
\end{tabular}%
}
\label{ab1}
\end{table}

\begin{table}[t]
\centering
\caption{Ablation results of the evaluation regimes. We replace the original ``8M Pretrain + 3M SFT'' validation with the evaluation approach followed in LLava-1.5 to prove the robustness of SAILViT.} 
\renewcommand{\arraystretch}{1}
\setlength{\tabcolsep}{5pt}
\resizebox{0.7\textwidth}{!}{%
\begin{tabular}{cccc}
\toprule
Visual Backbone Series & Freeze ViT in SFT & OpenSource & OpenCompass \\ \midrule
InternViT-300M-v2.5 & \ding{51} & 41.79 & 36.40 \\
AIMv2-Large & \ding{51} & 43.19 & 37.20 \\
\textbf{SAILViT-Large} & \ding{51} & \textbf{45.02} & \textbf{38.00} \\ \midrule
InternViT-300M-v2.5 & \ding{55} & 43.32 & 36.70 \\
AIMv2-Large & \ding{55} & 43.72 & 37.20 \\
\textbf{SAILViT-Large} & \ding{55} & \textbf{45.79} & \textbf{39.00} \\ \bottomrule
\end{tabular}%
}
\label{tab:sft}
\vspace{-2pt}
\end{table}

\vspace{-5pt}
\subsection{Ablation Study}
\label{sec: ablation study}
\vspace{-5pt}

In Table~\ref{ab1}, we implement four settings across three ablation configurations to explore the impact of the world knowledge infusion stage of the model training. In this section, we evaluate ViTs integrated with Qwen2.5-1.5B-based MLLMs, except for the experimental results in Table~\ref{abl:backone}, which used the InternLM2.5-1.8B. Detail results are shown in the Appendix~\ref{sec: ablation experiment details}.

\textbf{Effect of Trainable Parameters}. This experiment aims to observe if the training cost can be reduced without affecting performance by freezing the LLM parameters. From the top half of Table~\ref{ab1}, we find that unlocking the LLM parameters allows the ViT to be trained more adequately (setting 1 \textit{vs}. best setting). This phenomenon is reasonable because the model has a better understanding of the visual representations, which guides the visual backbone to produce more adaptive feature semantics.

\textbf{Scale of Integrated LLMs}. Here, we replace Qwen2.5-1.5B with the 0.5B version and perform the same three training stages of gradual feature refinement. Although the small-scale LLM can reduce the training cost appropriately, the ViT results obtained by integrating Qwen2.5-1.5B training are better (setting 2 \textit{vs}. best setting). We believe that the increasing number of LLM parameters contributes to more accurate cross-modal associations learned based on robust visual feature fusion.

\textbf{Rationality of Training Strategy}. We consider two patterns of Supervised Fine-Tuning (SFT) to infuse 36.5M world knowledge in the third training stage. The first one is to progressively perform SFT in multiple stages based on data quality differences, which is referred to as step-by-step. The second one is to implement a multi-task uniform SFT training in an All In One (AIO) manner. The results in Table~\ref{ab1} show that the AIO training achieves more competitive results on both benchmarks (setting 3 \textit{vs}. best setting). The potential explanation is that step-by-step training limits the data richness, which causes a learning bias of the model on the data distribution. In contrast, the AIO manner ensures more integrated knowledge learning and avoids the risk of model forgetting.

\textbf{Effectiveness of Evaluation Regimes}. To verify the robustness of SAILViT to different evaluation schemes when integrated in MLLMs, we replace the default evaluation settings in Section~\ref{sec:eval} with an evaluation pattern that follows the LLava-1.5~\cite{liu2024improved}.
Specifically, we first unlock the MLP for pre-training with 558K data. Then, we use 665k of mixed data for SFT training by unlocking MLP and LLM. The parameters of ViT are always frozen under the above settings. The results in the top half of Table~\ref{tab:sft} show that SAILViT-Large still outperforms AIMv2-Large and InternViT-300M-v2.5 on both benchmarks. In addition, we find that different visual backbones are capable of giving the model further performance improvements when all parameters are unlocked during the SFT training stage. This observation is reasonable since cross-modal feature alignment is deeply optimized. In summary, the best results achieved by SAILViT confirm the robustness and superiority of our method.

\begin{table}[t]
\centering
\caption{Ablation results of SAILViT-InternViT under diffenent evaluation regimes. To validate the effectiveness of training strategies of visual backbones, we train the SAILViT-InternViT based on InternViT and conduct experiments under two different evaluation regimes.}
\renewcommand{\arraystretch}{1.6}
\setlength{\tabcolsep}{2pt}
\resizebox{\textwidth}{!}{%
\begin{tabular}{ccccccccccc}
\toprule
\multicolumn{2}{c|}{Base Models} & \multicolumn{1}{c|}{\multirow{2}{*}{Avg.}} & \multirow{2}{*}{\makecell[c]{AI2D \\ (test)}} & \multirow{2}{*}{HallusionBench} & \multirow{2}{*}{\makecell[c]{MMBench \\ (val)}} & \multirow{2}{*}{\makecell[c]{MMMU \\ (val)}} & \multirow{2}{*}{MMVet} & \multirow{2}{*}{OCRBench} & \multirow{2}{*}{MMStar} & \multirow{2}{*}{\makecell[c]{MathVista \\ (testmini)}} \\ \cline{1-2}
LLM & \multicolumn{1}{c|}{Visual Backbones Series} & \multicolumn{1}{c|}{} &  &  &  &  &  &  &  &  \\ \midrule
\multicolumn{11}{c}{``8M Pretrain + 3M SFT'' Ablation Group} \\ \midrule
\multicolumn{1}{c|}{\multirow{2}{*}{Internlm2.5-1.8B}} & \multicolumn{1}{c|}{InternViT-300M-v2.5} & \multicolumn{1}{c|}{49.9} & 69.27 & \textbf{33.34} & 65.98 & 35.11 & 29.59 & 713 & 47.80 & 47.1 \\
\multicolumn{1}{c|}{} & \multicolumn{1}{c|}{\textbf{SAILViT-InternViT}} & \multicolumn{1}{c|}{\textbf{51.8}} & \textbf{70.73} & 33.23 & \textbf{67.72} & \textbf{35.33} & \textbf{32.25} & \textbf{746} & \textbf{50.87} & \textbf{49.3} \\ \midrule
\multicolumn{11}{c}{``LLava 558k Pretrain + 665k SFT'' Ablation Group} \\ \midrule
\multicolumn{1}{c|}{\multirow{2}{*}{Internlm2.5-1.8B}} & \multicolumn{1}{c|}{InternViT-300M-v2.5} & \multicolumn{1}{c|}{36.7} & 49.64 & \textbf{25.44} & 61.57 & \textbf{32.00} & \textbf{26.51} & 364 & 36.67 & 25.4 \\
\multicolumn{1}{c|}{} & \multicolumn{1}{c|}{\textbf{SAILViT-InternViT}} & \multicolumn{1}{c|}{\textbf{38.2}} & \textbf{52.33} & 25.41 & \textbf{63.47} & 30.11 & 24.04 & \textbf{410} & \textbf{41.67} & \textbf{27.4} \\ \bottomrule
\end{tabular}%
}
\label{abl:backone}
\vspace{-2pt}
\end{table}

\textbf{Scalability of Visual Backbones}. We explore generalizing the proposed gradual feature refinement strategy to other vision foundation models on the OpenCompass benchmark. Specifically, we choose InternViT-300M-v2.5 to perform the three-stage training procedure integrated with Qwen2.5-1.5B, resulting in the SAILViT-InternViT version. As shown in Table~\ref{abl:backone}, SAILViT-InternViT is superior to the baseline InternViT-300M-v2.5 in both ``8M Pretrain + 3M SFT'' and ``LLava 558k Pretrain + 665k SFT'' ablation groups, further confirming the scalability and generalizability of our strategy for different visual backbones.

In the Appendix~\ref{sec:data_abl}, we further provide the ablation study of data composition in the world knowledge infusion phase.

\vspace{-3pt}
\section{Comparison Results on Visual Recognition Tasks}
\vspace{-3pt}

\begin{table}[t]
\centering
\caption{Comparison results of different visual backbones on visual recognition tasks.}
\setlength{\tabcolsep}{2pt}
\resizebox{\textwidth}{!}{%
\begin{tabular}{c|cccccc}
\toprule
Settings    & \multicolumn{1}{l}{AIMv2-Large} & \begin{tabular}[c]{@{}c@{}}InternViT-300M-\\ 448px-V2.5\end{tabular} & \multicolumn{1}{l}{SAILViT-Large} & \multicolumn{1}{l}{AIMv2-Huge} & \multicolumn{1}{l}{SAILViT-Huge} & \begin{tabular}[c]{@{}c@{}}InternViT-6B-\\ 448px-V2.5\end{tabular} \\ \midrule
ImageNet-1k & 79.88\%                         & 73.70\%                                                              & 80.71\%                           & 81.68\%                        & 82.21\%                          & 84.18\%                                                            \\
ImageNet-A  & 25.41\%                         & 13.45\%                                                              & 29.31\%                           & 28.96\%                        & 33.04\%                          & 46.27\%                                                            \\
ImageNet-R  & 55.73\%                         & 39.99\%                                                              & 56.42\%                           & 58.13\%                        & 60.33\%                          & 59.88\%                                                            \\
ImageNet-V2 & 76.45\%                         & 69.55\%                                                              & 77.06\%                           & 77.32\%                        & 78.94\%                          & 80.92\%                                                            \\
Average     & 59.37\%                         & 49.17\%                                                              & 60.87\%                           & 61.52\%                        & 63.63\%                          & 67.81\%                                                            \\ \bottomrule
\end{tabular}
}
\label{vrt}
\end{table}

\begin{table}[t]
\centering
\caption{Model performance on visual tasks with different training magnitude data.}
\setlength{\tabcolsep}{15pt}
\resizebox{\textwidth}{!}{%
\begin{tabular}{cccccc}
\toprule
\multicolumn{1}{c|}{Settings}    & Origin  & proaio2m & proaio8m & proaio16m & proaio36.5m \\ \midrule
\multicolumn{6}{c}{SAILViT-Large}                                                          \\ \midrule
\multicolumn{1}{c|}{ImageNet-1k} & 79.88\% & 79.57\%  & 79.77\%  & 80.46\%   & 80.71\%     \\
\multicolumn{1}{c|}{ImageNet-A}  & 25.41\% & 27.72\%  & 26.68\%  & 28.07\%   & 29.31\%     \\
\multicolumn{1}{c|}{ImageNet-R}  & 55.73\% & 49.24\%  & 55.32\%  & 54.58\%   & 56.42\%     \\
\multicolumn{1}{c|}{ImageNet-V2} & 76.45\% & 75.31\%  & 76.12\%  & 77.03\%   & 77.06\%     \\
\multicolumn{1}{c|}{Average}     & 59.37\% & 57.96\%  & 59.47\%  & 60.03\%   & 60.87\%     \\ \midrule
\multicolumn{6}{c}{SAILViT-Huge}                                                          \\ \midrule
\multicolumn{1}{c|}{ImageNet-1k} & 81.68\% & 81.09\%  & 81.52\%  & 81.96\%   & 82.21\%     \\
\multicolumn{1}{c|}{ImageNet-A}  & 28.96\% & 32.52\%  & 35.51\%  & 32.01\%   & 33.04\%     \\
\multicolumn{1}{c|}{ImageNet-R}  & 58.13\% & 54.85\%  & 57.24\%  & 56.01\%   & 60.33\%     \\
\multicolumn{1}{c|}{ImageNet-V2} & 77.32\% & 77.12\%  & 78.01\%  & 78.28\%   & 78.94\%     \\
\multicolumn{1}{c|}{Average}     & 61.52\% & 61.40\%  & 63.07\%  & 62.07\%   & 63.63\%     \\ \toprule
\end{tabular}
}
\label{table:scaling_imagenet}
\end{table}

\begin{figure}[t]
  \centering
  \includegraphics[width=0.95\linewidth]{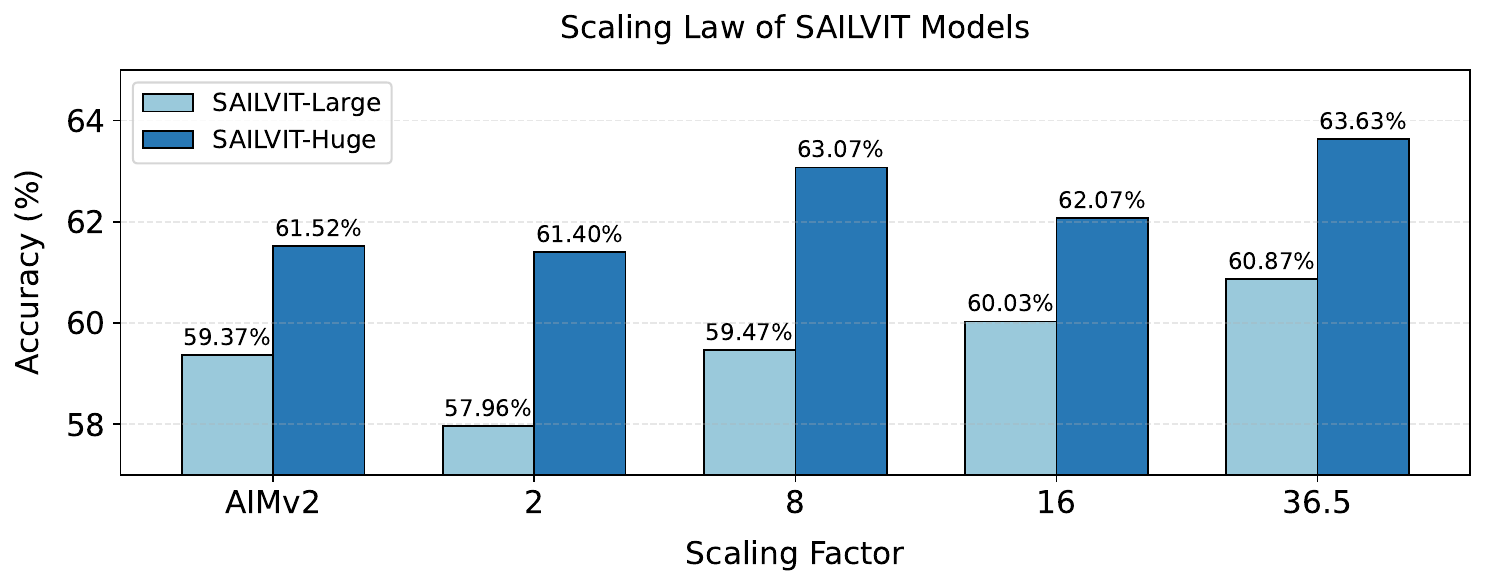}
  \caption{Scaling trend of average gains for SAILViT-Large/Huge.}
  \label{fig:scaling_imagenet}
  \vspace{-6pt}
\end{figure}

Further, we explore the performance of SAILViT on visual recognition tasks. To maintain the fair experimental regime, we feed the output of the token from the last block of the different visual backbones into the two-layer MLP after averaging them,\textit{ i.e.}, a token representing the probability value is finally outputted. All models are trained for 10 epochs on the ImageNet-1k  training set and then evaluated on the ImageNet-1k~\cite{5206848}, ImageNet-A~\cite{hendrycks2021naturaladversarialexamples}, ImageNet-R~\cite{hendrycks2021facesrobustnesscriticalanalysis}, and ImageNet-v2~\cite{recht2019imagenetclassifiersgeneralizeimagenet} test sets.
Table~\ref{vrt} shows the comprehensive comparison results. SAILViT-Large improves on average by 1.5\% compared to baseline AIMv2-Large, while significantly improving by 11.7\% compared to InternVIT-300M. Our model shows significant gains on all test sets. Similarly, SAILViT-Huge improves the results of AIMv2-Large regarding 2.11\% on average. The proposed model exhibits comparable performance to InternViT-6B, despite having fewer parameters, and yields better results on ImageNet-R. The above experimental results demonstrate that the ViT obtained through the proposed three-stage progressive training approach is capable of extracting more effective visual features and consistently outperforms other models on classification tasks in visual perception.

Table~\ref{table:scaling_imagenet} presents the performances of models trained on visual tasks using different training magnitude data in the third training phase, \textit{i.e.}, the world knowledge infusion phase. We find that as the training data magnitude increases, the model performs better on visual recognition tasks. Figure~\ref{fig:scaling_imagenet} intuitively shows the scaling plots of average gains for the Large and Huge versions of SAILViT. We observe that there is still room for continued improvement in the SAILViT performance, further demonstrating the effectiveness of our approach.

\vspace{-3pt}
\section{Conclusion and Discussion}
\vspace{-3pt}
In this work, we propose a three-stage training pipeline for visual backbones based on gradual feature refinement, aiming to bridge the feature space gap between visual features and LLM features caused by independent pretraining among MLLMs. Through this training approach, we develop two model-scale visual backbones specifically tailored for MLLMs: SAILViT-Large and SAILViT-Huge. Our experiments demonstrate that SAILViT-based MLLMs, when combined with different model-scaled LLMs from various families, consistently outperform those using other open-source ViTs such as AIMv2-Large/Huge and InternViT. This not only validates the effectiveness, robustness, and generalizability of SAILViT but also highlights its superiority in feature alignment with LLMs. Furthermore, we train SAILViT-InternViT based on InternViT, and experimental results show that it exhibits stronger affinity with LLMs compared to the original InternViT, further verifying the generalizability of our proposed training method.
\paragraph{Limitations.} The proposed SAILViT demonstrates excellent effectiveness, generalizability, and robustness through the evaluation regimes we designed with reference to LLava-1.5. However, the training strategies of mainstream MLLMs are often more complex and visual backbone-friendly. For instance, Ovis2 adopts a progressive alignment training strategy, which partially bridges the gap between the original feature spaces of ViT and LLMs. Although we have validated the robustness of SAILViT within the evaluation regimes incorporating the ViT progressive alignment stage in the Appendix~\ref{sec: futher verification SAILViT robustness}, whether SAILViT can still maintain its advantages over other visual backbones under larger dataset scales and more complex training strategies requires further investigation. Additionally, current top-performance open-source MLLMs often employ ViTs with larger parameters. For example, InternVL3-38B/78B utilizes InternViT-6B, and Ovis2-34B employs AIMv2-1B to extract visual features. Whether our proposed SAILViT can maintain advantages compared to these ViTs with larger parameter scales still requires further exploration.
\paragraph{Future Work.} In future works, we will first explore whether a three-stage gradual feature refinement training strategy can compress effective visual information into fewer tokens, enabling the modeling of longer visual feature inputs. Moreover, we intend to conduct training on SAILViT with substantially larger parameter magnitudes, leveraging open-source ViTs, such as AIMv2-1B/3B and InternViT-6B, to systematically explore the upper bounds of parameter scaling for SAILViT. Finally, we aim to investigate the construction of ViT supporting native resolution, avoiding the loss of detailed visual information caused by image slicing and facilitating the modeling of finer-grained visual features. 
\paragraph{Broader Impacts.}
We provide more discussions of the broader impacts in the Appendix~\ref{sec: broader impacts}.

\bibliographystyle{plain}
\bibliography{arxiv}

\newpage
\appendix

\section{Further Verification of the Robustness of SAILViT} 
\label{sec: futher verification SAILViT robustness}
Considering that current top-performance MLLMs employ more complex training strategies, often involving gradual alignment training stages for visual backbones, such approaches are more friendly to ViTs and can partially bridge the feature space gap caused by parameter misalignment between pretrained ViTs and LLMs due to independent training. Under such a granular training phase design, the advantages of SAILViT may be diluted. To further validate the robustness of SAILViT, we introduce a gradual alignment training stage for ViTs into the existing ViT evaluation regimes. Specifically, we insert a new training stage into the original two-stage training paradigm, where we simultaneously unfroze the parameters of both the connector and ViT while freezing the LLM, using a learning rate of 2e-5 and a batch size of 512. As shown in Table~\ref{tab: three stege abl}, the additional ViT training stage indeed improved the overall performance of MLLMs, and compared to the baseline without this stage, the model consistently achieved higher scores on the OpenCompass benchmark, demonstrating the benefit of gradual parameter alignment for ViTs. Experimental results show that even when attempting to weaken the gap from ViT's original pretrained parameters by adding dedicated ViT training stages, SAILViT still outperforms other open-source ViTs, further confirming its robustness.

\begin{table}[h]
\centering
\caption{Ablation results under different evaluation regimes. We introduce a new ViT-friendly training stage into the original two-stage training paradigm, establishing a three-stage training framework. The resulting MLLMs are systematically evaluated on the OpenCompass benchmark.}
\renewcommand{\arraystretch}{1.6}
\setlength{\tabcolsep}{2pt}
\resizebox{0.6\textwidth}{!}{%
\begin{tabular}{cccc}
\toprule
Visual Backbone Series & LLM & \makecell[c]{Two-stage \\ Training} & \makecell[c]{Three-stage \\ Training} \\ \midrule
\multicolumn{1}{c|}{AIMv2-Large} & \multicolumn{1}{c|}{\multirow{4}{*}{Qwen2.5-1.5B}} & 54.6 & 55.1 \\
\multicolumn{1}{c|}{\textbf{SAILViT-Large}} & \multicolumn{1}{c|}{} & \textbf{56.9} & \textbf{56.9} \\ \cline{1-1} \cline{3-4} 
\multicolumn{1}{c|}{AIMv2-Huge} & \multicolumn{1}{c|}{} & 56.3 & 56.6 \\
\multicolumn{1}{c|}{\textbf{SAILViT-Huge}} & \multicolumn{1}{c|}{} & \textbf{57.7} & \textbf{58.0} \\ \bottomrule
\end{tabular}%
}
\label{tab: three stege abl}
\end{table}

\section{Fine-grained Comparison Results on the OpenSource}
\label{sec:a}

\begin{table}[h]
\caption{
Comparison results between SAILViT and other open-source ViTs on the OpenSource benchmark when connecting to different LLM series with different parameters. The best results are bolded on the quantitative metrics. OCRVQA$^\ast$ means the OCRVQA\_TESTCORE.}
\resizebox{\textwidth}{!}{%
\renewcommand{\arraystretch}{1.4}
\setlength{\tabcolsep}{1pt}
\begin{tabular}{c|cc|c|ccccccccccccccccccc}
\toprule
\multicolumn{3}{c|}{Base Models} & \multirow{2}{*}{Avg.} & \multirow{2}{*}{\makecell[c]{AI2D \\ (test)}} & \multirow{2}{*}{\makecell[c]{ChartQA \\ (test)}} & \multirow{2}{*}{\makecell[c]{DocVQA \\ (val)}} & \multirow{2}{*}{\makecell[c]{Hallusion \\ Bench}} & \multirow{2}{*}{\makecell[c]{InfoVQA \\ (val)}} & \multirow{2}{*}{\makecell[c]{LLaVA \\ Bench}} & \multirow{2}{*}{\makecell[c]{MMBench \\ (val)}} & \multirow{2}{*}{MME} & \multirow{2}{*}{\makecell[c]{MMMU \\ (val)}} & \multirow{2}{*}{MMStar} & \multirow{2}{*}{MMVet} & \multirow{2}{*}{\makecell[c]{MathVista \\ (testmini)}} & \multirow{2}{*}{\makecell[c]{OCR \\ Bench}} & \multirow{2}{*}{OCRVQA$^\ast$} & \multirow{2}{*}{POPE} & \multirow{2}{*}{\makecell[c]{Real \\ WorldQA}} & \multirow{2}{*}{\makecell[c]{SEED \\ Bench\_IMG}} & \multirow{2}{*}{\makecell[c]{ScienceQA \\ (val)}} & \multirow{2}{*}{\makecell[c]{TextVQA \\ (val)}} \\ \cline{1-3}
\multicolumn{1}{c}{LLM Series} & \multicolumn{2}{c|}{Visual Backbones} &  &  &  &  &  &  &  &  &  \\ 
\midrule
\rowcolor{gray!15} \multicolumn{23}{c}{\textit{InternLM2.5 Series}} \\ 
\midrule
\multirow{5}{*}{InternLM2.5-1.8B} & InternViT-300M-v2.5 & \multicolumn{1}{l|}{\multirow{3}{*}{300M}} & 56.74 & 69.27 & 67.08 & 73.90 & 33.34 & 47.84 & 34.70 & 65.98 & 1637 & 35.11 & 47.80 & 29.59 & 47.10 & 713 & 34.47 & 87.76 & 56.08 & 67.90 & 91.46 & 58.92 \\
\multicolumn{1}{l|}{} & Aimv2-large & & 57.34 & 71.53 & 66.32 & 73.06 & 33.02 & 48.53 & 37.50 & 66.45 & 1641 & 35.33 & 49.93 & 32.34 & 48.00 & \textbf{729} & \textbf{35.35} & 88.33 & 58.30 & 70.11 & 91.03 & 52.82 \\
\multicolumn{1}{l|}{} & \textbf{SailViT-300M} & & \textbf{59.09} & \textbf{73.09} & \textbf{69.24} & \textbf{75.32} & \textbf{35.32} & \textbf{50.01} & \textbf{39.70} & \textbf{67.80} & \textbf{1750} & \textbf{36.78} & \textbf{50.80} & \textbf{34.40} & \textbf{49.30} & 716 & 35.12 & \textbf{88.94} & \textbf{60.13} & \textbf{70.80} & \textbf{91.75} & \textbf{60.11} \\
 \cline{2-23}
 & AIMv2-huge & \multicolumn{1}{l|}{\multirow{2}{*}{600M}} & 58.38 & 72.44 & 68.00 & 74.57 & 29.84 & 50.50 & 38.00 & 68.34 & \textbf{1668} & 33.11 & 51.47 & 34.40 & 49.70 & \textbf{759} & 34.96 & \textbf{89.78} & 58.04 & 70.73 & 93.09 & 56.87 \\
\multicolumn{1}{l|}{} & \textbf{SailViT-600M} & & \textbf{60.48} & \textbf{73.19} & \textbf{69.96} & \textbf{76.94} & \textbf{37.62} & \textbf{52.20} & \textbf{42.20} & \textbf{70.16} & 1640 & \textbf{36.89} & \textbf{53.20} & \textbf{37.16} & \textbf{51.20} & 757 & \textbf{35.87} & 89.59 & \textbf{60.92} & \textbf{71.58} & \textbf{93.47} & \textbf{62.64} \\ 
\midrule
\rowcolor{gray!15} \multicolumn{23}{c}{\textit{Qwen2.5 Series}} \\ 
\midrule
\multirow{5}{*}{Qwen2.5-1.5B} & InternViT-300M-v2.5 & \multicolumn{1}{l|}{\multirow{3}{*}{300M}} & 59.09 & 64.02 & 67.24 & 77.09 & 33.56 & 46.93 & 43.00 & 69.16 & 1692 & 41.56 & 48.33 & 38.72 & 50.00 & 725 & 35.58 & 88.41 & 54.64 & 69.11 & 92.27 & \textbf{70.10} \\
 & Aimv2-large & & 60.05 & 75.29 & 68.84 & 78.56 & 33.73 & 56.93 & \textbf{45.00} & 70.94 & 1754 & 40.56 & 53.33 & 37.16 & 50.80 & 749 & 35.35 & 88.29 & 57.39 & 70.97 & 92.61 & 47.67 \\
 & \textbf{SailViT-300M} & & \textbf{62.46} & \textbf{76.20} & \textbf{72.16} & \textbf{81.06} & \textbf{40.93} & \textbf{60.19} & 44.50 & \textbf{71.94} & \textbf{1903} & \textbf{41.89} & \textbf{53.80} & \textbf{38.99} & \textbf{53.90} & \textbf{772} & \textbf{36.78} & \textbf{88.69} & \textbf{60.13} & \textbf{71.68} & \textbf{94.13} & 54.66 \\
 \cline{2-23}
 & AIMv2-huge & \multicolumn{1}{l|}{\multirow{2}{*}{600M}} & 61.07 & 77.66 & \textbf{73.08} & 79.58 & 35.44 & 59.53 & \textbf{43.20} & 72.33 & 1714 & 42.56 & 54.13 & 39.72 & 51.40 & 769 & \textbf{37.27} & 88.76 & 58.82 & 71.25 & 93.94 & 43.52 \\
 & \textbf{SailViT-600M} & & \textbf{62.72} & \textbf{78.27} & 72.32 & \textbf{82.14} & \textbf{37.91} & \textbf{60.84} & 40.50 & \textbf{73.99} & \textbf{1927} & \textbf{43.44} & \textbf{55.60} & \textbf{40.55} & \textbf{52.10} & \textbf{795} & 36.85 & \textbf{89.69} & \textbf{60.78} & \textbf{71.73} & \textbf{94.09} & \textbf{52.52} \\ \midrule
\multirow{5}{*}{Qwen2.5-7B} & InternViT-300M-v2.5 & \multicolumn{1}{l|}{\multirow{3}{*}{300M}} & 67.17 & 81.28 & 74.48 & 84.43 & 44.82 & 66.46 & 49.50 & 77.67 & 2114 & 49.44 & 59.27 & 43.90 & 62.20 & 784 & 37.24 & 86.59 & 61.70 & 74.10 & 96.57 & 72.73 \\
 & Aimv2-large & & 68.98 & 81.74 & 74.04 & 86.47 & 45.36 & 68.61 & 55.00 & 77.79 & 2208 & 49.00 & 60.80 & 48.07 & 64.00 & \textbf{828} & 37.27 & 87.83 & \textbf{65.10} & 75.64 & \textbf{97.52} & 74.73 \\
 & \textbf{SailViT-300M} & & \textbf{70.04} & \textbf{82.12} & \textbf{78.80} & \textbf{87.05} & \textbf{45.63} & \textbf{69.11} & \textbf{55.20} & \textbf{78.95} & \textbf{2288} & \textbf{51.67} & \textbf{60.87} & \textbf{49.50} & \textbf{67.10} & 805 & \textbf{38.35} & \textbf{88.32} & 64.84 & \textbf{75.91} & 96.90 & \textbf{78.32} \\
 \cline{2-23}
 & AIMv2-huge & \multicolumn{1}{l|}{\multirow{2}{*}{600M}} & 69.71 & 81.44 & 79.00 & 87.11 & 44.04 & 69.44 & 51.30 & \textbf{80.30} & 2181 & \textbf{50.78} & 62.33 & 46.10 & \textbf{67.40} & 815 & 37.37 & 87.46 & 68.76 & \textbf{76.44} & \textbf{97.76} & 78.09 \\
 & \textbf{SailViT-600M} & & \textbf{70.90} & \textbf{83.00} & \textbf{79.48} & \textbf{88.58} & \textbf{48.65} & \textbf{70.65} & \textbf{56.70} & 79.64 & \textbf{2238} & 50.33 & \textbf{62.60} & \textbf{49.22} & 65.20 & \textbf{833} & \textbf{38.09} & \textbf{88.35} & \textbf{69.28} & 76.36 & 97.19 & \textbf{80.47} \\ 
\midrule
\rowcolor{gray!15} \multicolumn{23}{c}{\textit{Qwen3 Series}} \\ 
\midrule
\multirow{2}{*}{Qwen3-0.6B} & AIMv2-large & \multicolumn{1}{l|}{\multirow{2}{*}{300M}} & 57.45 & \textbf{71.37} & 67.08 & 74.03 & 38.92 & 46.89 & \textbf{41.30} & \textbf{65.52} & 1613 & 35.11 & 49.80 & 33.07 & 49.20 & 703 & 33.92 & 85.03 & 57.25 & 70.10 & 93.85 & 51.15 \\
 & \textbf{SailViT-300M} & & \textbf{59.64} & 71.05 & \textbf{69.60} & \textbf{76.44} & \textbf{41.05} & \textbf{51.31} & 40.30 & 64.86 & \textbf{1683} & \textbf{36.67} & \textbf{50.93} & \textbf{34.77} & \textbf{49.40} & \textbf{741} & \textbf{35.58} & \textbf{86.67} & \textbf{60.92} & \textbf{70.50} & \textbf{95.14} & \textbf{63.86} \\ \midrule
\multirow{4}{*}{Qwen3-1.7B} & AIMv2-large & \multicolumn{1}{l|}{\multirow{2}{*}{300M}} & 62.03 & 77.49 & 73.36 & 80.97 & 39.76 & 59.10 & 39.30 & 71.28 & \textbf{1894} & \textbf{42.78} & 53.40 & 39.17 & 51.40 & 751 & 36.72 & \textbf{86.28} & 58.30 & 72.58 & 96.19 & 57.69 \\
 & \textbf{SailViT-300M} & & \textbf{64.47} & \textbf{78.82} & \textbf{75.52} & \textbf{82.73} & \textbf{41.84} & \textbf{60.99} & \textbf{47.50} & \textbf{71.44} & 1892 & 42.56 & \textbf{54.80} & \textbf{43.21} & \textbf{52.80} & \textbf{790} & \textbf{37.60} & 85.61 & \textbf{61.31} & \textbf{73.17} & \textbf{96.38} & \textbf{72.18} \\
 \cline{2-23}
 & AIMv2-huge & \multicolumn{1}{l|}{\multirow{2}{*}{600M}} & 63.19 & 79.24 & 66.80 & 81.41 & 40.44 & 60.47 & 45.50 & 71.09 & \textbf{1962} & 42.00 & 54.80 & 43.07 & 52.90 & 781 & 36.95 & \textbf{86.67} & 60.13 & 73.17 & \textbf{96.71} & 61.13 \\
 & \textbf{SailViT-600M} & & \textbf{65.86} & \textbf{79.89} & \textbf{76.88} & \textbf{83.23} & \textbf{41.23} & \textbf{63.18} & \textbf{49.50} & \textbf{74.19} & 1917 & \textbf{43.11} & \textbf{56.93} & \textbf{44.31} & \textbf{54.90} & \textbf{806} & \textbf{37.83} & 86.26 & \textbf{64.71} & \textbf{74.19} & 96.33 & \textbf{75.53} \\ \midrule
\multirow{2}{*}{Qwen3-8B} & AIMv2-huge & \multicolumn{1}{l|}{\multirow{2}{*}{600M}} & 69.60 & 83.35 & 43.68 & 88.53 & 43.50 & 74.91 & \textbf{63.30} & 80.92 & 2135 & \textbf{52.78} & \textbf{65.60} & \textbf{50.05} & 67.80 & 839 & 38.22 & 89.30 & 68.63 & 76.48 & 97.42 & 77.81 \\
 & \textbf{SailViT-600M} & & \textbf{70.58} & \textbf{84.16} & \textbf{52.60} & \textbf{89.69} & \textbf{48.23} & \textbf{76.31} & 57.50 & \textbf{81.89} & \textbf{2230} & 51.78 & 65.13 & 46.83 & \textbf{69.40} & \textbf{857} & \textbf{38.28} & \textbf{89.30} & \textbf{69.41} & \textbf{76.85} & \textbf{97.95} & \textbf{80.28} \\ \bottomrule
\end{tabular}
}
\label{aupp_opensource}
\end{table}

To further demonstrate the robustness and generalizability of SAILViT, we show the quantitative results at a fine-grained level for all datasets on the OpenSource benchmark in Table~\ref{aupp_opensource}. (i) We find that both SAILViT-Large and Huge versions exhibit significant performance improvements on the vast majority of datasets when paired with different families of LLMs, implying that SAILViT can help different MLLMs further break through the potential performance bottlenecks due to modality discrepancies and target training parameter conflicts.
 (ii) SAILViT further provides consistent performance gains when paired with different-sized LLMs of the same series. Similarly, the Huge version of the proposed visual backbone also achieves better results overall compared to the Large version. These observations demonstrate that our gradual feature
refinement training strategy can further strengthen the gains of the parameter scaling law, providing new insights into the design and selection of vision foundation models in the subsequent MLLM construction. (iii) Other cross-dataset observations align the phenomena found in the main paper, verifying the effectiveness of SAILViT on different downstream multimodal perception tasks.

\section{Scaling Law of SAILViT-Large with Qwen2.5-0.5B}
\label{sec: scaling law under 0d5B}
\begin{figure}[h]
  \centering
  \includegraphics[width=0.8\linewidth]{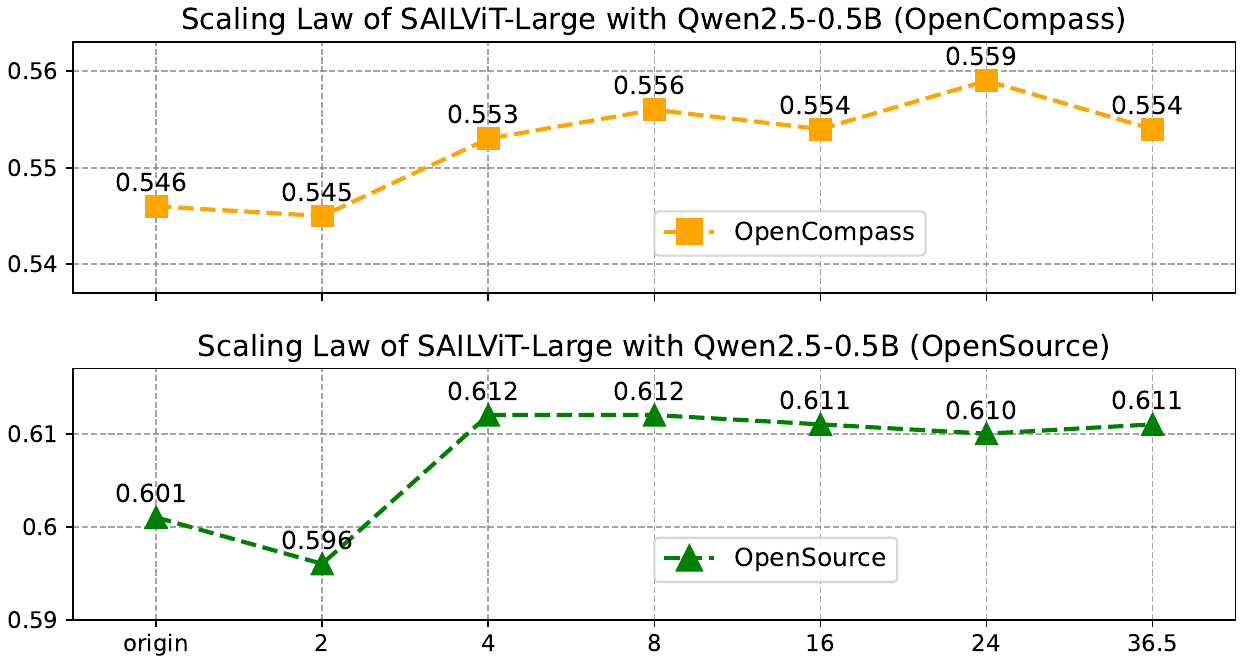}
  \caption{Illustration of the scaling law phenomenon analysis. The overall performances of MLLMs paired with SAILViT-Large and Qwen2.5-0.5B on OpenCompass and OpenSource benchmarks show increasing trends as the amount of data in the world knowledge infusion stage increases. ``origin'' means the results of the MLLM equipped with AIMv2-Large.}
  \label{fig:0d5 scaling}
  \vspace{-6pt}
\end{figure}

We replace the LLM adapted in the three-stage training of ViT from Qwen2.5-1.5B with Qwen2.5-0.5B. Notably, a consistent trend is observed in Figure~\ref{fig:0d5 scaling}: as the data volume in the world knowledge infusion stage increased, the performance of SAILViT correspondingly improved.

\section{Abalation Experiment Details}
Here, we present the detailed information of the ablation experiments in Section~\ref{sec: ablation study}.
\label{sec: ablation experiment details}

\begin{table}[h]
\centering
\caption{Detail ablation results on SAILViT in the world knowledge infusion stage.}
\setlength{\tabcolsep}{2pt}
\resizebox{0.8\textwidth}{!}{%
\begin{tabular}{cccc}
\toprule
\textbf{Ablation} & \textbf{Training Details} & \textbf{OpenSource} & \textbf{OpenCompass} \\ \midrule
Baseline & Normal & 60.05 & 54.6 \\ \midrule
Freeze LLM & World knowledge infusion & 62.45 & 55.3 \\ \midrule
\multirow{6}{*}{Integrated with Qwen2.5-0.5B} & AIO SFT 2M & 59.56 & 54.5 \\
 & AIO SFT 4M & 61.17 & 55.3 \\
 & AIO SFT 8M & 61.22 & 55.6 \\
 & AIO SFT 16M & 61.06 & 55.4 \\
 & AIO SFT 24M & 61.02 & 55.9 \\
 & AIO SFT 36.5M & 61.10 & 55.4 \\ \midrule
\multirow{3}{*}{Step by Step} & First Stage 21M & 61.43 & 55.4 \\
 & Second Stage 12M & 62.08 & 56.2 \\
 & Third Stage 3.5M & 61.82 & 55.8 \\ \midrule
\multirow{6}{*}{AIO} & AIO SFT 2M & 61.74 & 55.5 \\
 & AIO SFT 4M & 60.94 & 55.7 \\
 & AIO SFT 8M & 62.10 & 56.4 \\
 & AIO SFT 16M & 61.81 & 55.9 \\
 & AIO SFT 24M & 61.60 & 56.0 \\
 & AIO SFT 36.5M & 62.46 & 56.9 \\ \bottomrule
\end{tabular}%
}
\end{table}

\section{Broader Impacts}
Our proposed SAILViT outperforms the ViTs utilized in current top-performance lightweight MLLMs (<10B), demonstrating significant superiority across OpenCompass and OpenSource benchmark. This showcases its remarkable effectiveness, generalization, and robustness, indicating its potential to elevate the performance ceiling of lightweight MLLMs. Additionally, the gradual feature refinement training approach for ViT introduced in this study exhibits excellent generalizability, enabling its seamless application across diverse visual backbones. This dual impact not only advances the state-of-the-art in lightweight MLLMs but also paves the way for broader applications in computer vision research and development.
\label{sec: broader impacts}

\section{Data Composition Ablation}
\label{sec:data_abl}

\begin{figure}[t]
  \centering
  \includegraphics[width=\linewidth]{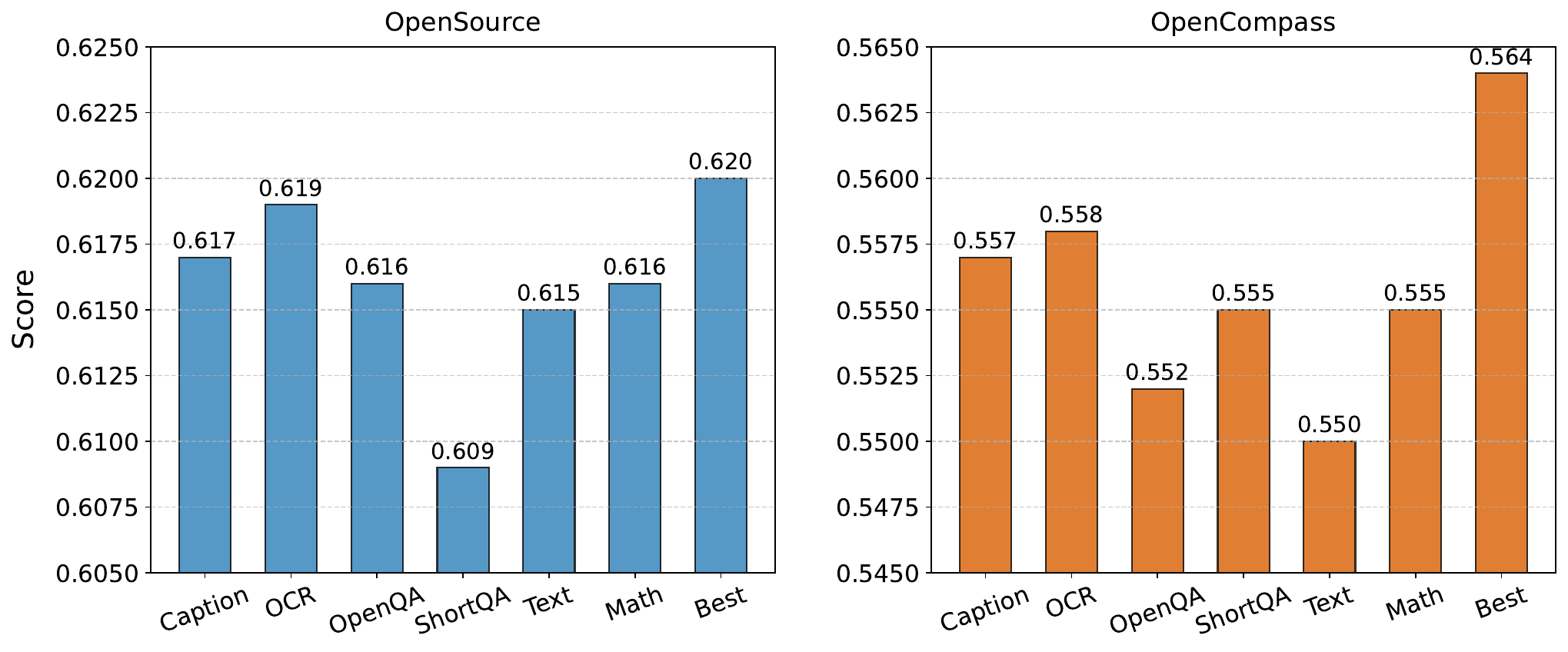}
  \caption{Ablation of data proportions in the world knowledge infusion stage. The total quantity of data remains consistent across different proportion configurations. The horizontal axis indicates the proportion of this major data category to 50\%, while the vertical axis represents the indicator results within the ablation framework. These results are derived from SAILViT models trained under varying World Knowledge Infusion data proportions, which are subsequently integrated into Qwen2.5-1.5B.}
  \label{supp_fig}
\end{figure}

To further investigate the impact of different data ratios during the world knowledge infusion stage on the performance of SAILViT, we first categorize the data at this stage into six major types: Caption, OCR, OpenQA, ShortQA, Text, and Math. To ensure the fairness of the experiments, we keep the total amount of data with different ratios fixed at 8M. We sequentially increase the quantity of each of these six types of data to a dominant position, \textit{i.e.}, 50\%, while the proportion of the remaining data types decreased overall, but their relative concentrations remained unchanged. The experiment corresponding to ``Best'' is the data ratio we actually adopted, which corresponds to the data ratio in Table 1 of the main paper. We respectively integrate the ViTs trained under these different data ratios into Qwen2.5-1.5B and evaluate them within the same ablation regimes. As shown in Figure~\ref{supp_fig}, excessively increasing the concentration ratio of any type of data will affect the performance of ViTs. These experimental phenomena provide valuable insights for the progressive training and research of SAILViT.


\end{document}